# *LATTE*: <u>L</u>STM Self-<u>Att</u>ention based Anomaly Detection in <u>E</u>mbedded Automotive Platforms

VIPIN KUMAR KUKKALA, Colorado State University

SOORYAA VIGNESH THIRULOGA, Colorado State University

SUDEEP PASRICHA, Colorado State University

Modern vehicles can be thought of as complex distributed embedded systems that run a variety of automotive applications with real-time constraints. Recent advances in the automotive industry towards greater autonomy are driving vehicles to be increasingly connected with various external systems (e.g., roadside beacons, other vehicles), which makes emerging vehicles highly vulnerable to cyber-attacks. Additionally, the increased complexity of automotive applications and the in-vehicle networks results in poor attack visibility, which makes detecting such attacks particularly challenging in automotive systems. In this work, we present a novel anomaly detection framework called *LATTE* to detect cyber-attacks in Controller Area Network (CAN) based networks within automotive platforms. Our proposed *LATTE* framework uses a stacked Long Short Term Memory (LSTM) predictor network with novel attention mechanisms to learn the normal operating behavior at design time. Subsequently, a novel detection scheme (also trained at design time) is used to detect various cyber-attacks (as anomalies) at runtime. We evaluate our proposed *LATTE* framework under different automotive attack scenarios and present a detailed comparison with the best-known prior works in this area, to demonstrate the potential of our approach.

CCS CONCEPTS • **Security and privacy~Network security** • **Networks~Network types~Cyber-physical networks** • **Computer systems organization~Embedded and cyber-physical systems~Embedded systems**

**Additional Keywords and Phrases:** Anomaly detection, automotive networks, machine learning, recurrent neural networks, cyber-physical systems

## 1 INTRODUCTION

Modern vehicles are experiencing a rapid increase in the complexity of embedded systems integrated into various vehicle subsystems, due to the increased interest in autonomous driving. The aggressive competition between automakers to reach autonomy goals is further driving the complexity of Electronic Control Units (ECUs) and the communication network that connects them [1]. Additionally, recent solutions for Advanced Driver Assistance Systems (ADAS) require interactions with various external systems using a variety of communication standards such as 5G, Wi-Fi, and Vehicle-to-X (V2X) protocols [2]. The V2X communication facilitates a spectrum of connections such as vehicle-to-vehicle (V2V), vehicle-to-pedestrian (V2P), vehicle-to-infrastructure (V2I), and vehicle-to-cloud (V2C) [3]. These new solutions are transforming modern vehicles by making them more connected to the external environment. To support the increasingly sophisticated ADAS functionality and connectivity to the outside world, highly complex software is required to run on the ECUs in such vehicles, to handle highly safety-critical and time-sensitive automotive applications, e.g., pedestrian and traffic sign detection, lane changing, automatic parking, path planning, etc. This increased software and hardware complexity of the automotive electrical/electronic (E/E) architecture and increased connectivity with external systems has an important implication: it provides a large attack surface and thus gives rise to more opportunities for attackers to gain unauthorized access to the in-vehicle network and execute cyber-attacks. The complexity in emerging vehicles also leads to poor attack visibility over the network, making it hard to detect attacks that can be easily hidden within normal operational activities. Such cyber-attacks on vehicles can induce various anomalies in the network, altering the normal behavior of the network as well as the compute system (ECU) behavior. Due to the time-sensitive and safety-critical nature of automotive applications, any minor instability in the system due to these induced anomalies could lead to a major catastrophe, e.g., delaying the perception of a pedestrian, preventing an airbag from deploying in the case of a collision, or erroneously changing lanes into oncoming traffic, due to maliciously corrupted sensor readings.

An attack via an externally-linked component or compromised ECU can manifest in several forms over the in-vehicle network. One of the most commonly observed attacks is flooding the in-vehicle network with random or specific messages which increases the overall network load and results in halting any useful activity over the network. An advanced remote attack on an ECU could involve sending a kill command to the engine during normal driving. More sophisticated attacks could involve installing malware on the ECU and using it to achieve malicious goals. Some of the recent state-of-the-art attacks have used a vehicle's infotainment system as an attack vector to launch buffer overflow and denial of service attacks [5], performed reverse engineering of keyless entry automotive systems to wirelessly lock pick the vehicle immobilizer [6], etc. Researchers in [4] foresee a much more severe attack involving potentially targeting the U. S. electric power grid by using public electric vehicle charging stations as an attack vector to infect vehicles that use these stations with malware. Many other attacks on real-world vehicles are presented in [7]-[10]. The common aspect of these attacks is that they involve gaining unauthorized access to the in-vehicle network and modifying certain fields in the message frames, thereby tricking the receiving ECU into thinking that the malicious message is legitimate. All of these attacks can have catastrophic effects and need to be detected before they are executed. This problem will get exacerbated with the onset of connected and autonomous vehicles. Hence, restricting external attackers via early detection of their attacks is vital to realizing secure automotive systems.

Conventional computer networks utilize protective mechanisms such as firewalls (software) and isolation units such as gateways and switches (hardware) to protect from external attacks [11]. However, persistent attackers have been coming up with advanced attacks that leverage the increased compute and communication capabilities in modern ECUs, causing the traditional protection systems to become obsolete. This raises a need for a system-level solution that can continuously monitor the vehicle network, to detect cyber-attacks. One promising solution is to deploy a software framework for anomaly detection, which involves monitoring the network for unusual activities and raising an alarm when suspicious activity is detected. This approach can be extended to detect and classify various types of attacks on the in-vehicle network. Such a framework can learn the normal system behavior at design time and monitor the network for anomalies at runtime. A traditional approach for anomaly detection uses rule-based approaches such as monitoring message frequency [12], memory heat map [13], etc., to detect known attack signatures. However, due to the increased complexity of cyber-attacks, such traditional rule-based systems fail to recognize new and complex patterns, rendering these approaches ineffective. Fortunately, recent advances in deep learning and the availability of in-vehicle network data have brought forth the possibility of using sophisticated deep learning models for anomaly detection.

In this work, we present a novel anomaly detection framework called *LATTE* to detect cyber-attacks in the Controller Area Network (CAN) based automotive networks. Our proposed *LATTE* framework uses sequence models in deep learning in an unsupervised setting to learn the normal system behavior. *LATTE* leverages that information at runtime to detect anomalies by observing for any deviations from the learned normal behavior. This is illustrated in Fig. 1. The plot on the top right shows the expected deviation (computed using the model that was trained at design time) vs the observed deviation. The divergence in signal values during the attack intervals (shown in red area) can be used as a metric to detect cyber-attacks as anomalies. Our proposed *LATTE* framework aims to maximize the anomaly detection accuracy, precision, and recall, while minimizing the false-positive rate. Our novel contributions in this work can be summarized as follows:

- We propose a stacked Long-Short Term Memory (LSTM) based predictor model that integrates a novel self-attention mechanism to learn the normal automotive system behavior at design time;
- We design a one class support vector machine (OCSVM) based detector that works with the LSTM self-attention predictor model to detect different cyber-attacks at runtime;
- We present modifications to existing vehicle communication controllers that can help in realizing the proposed anomaly detection system on a real-world ECU;
- We perform a comprehensive analysis on the selection of deviation measures that quantify the deviation from the normal system behavior;
- We explore several variants of our proposed *LATTE* framework and selected the best performing one, which is then compared with the best-known prior works in the area (statistical-based, proximity-based, and ML-based works), to show *LATTE*'s effectiveness.

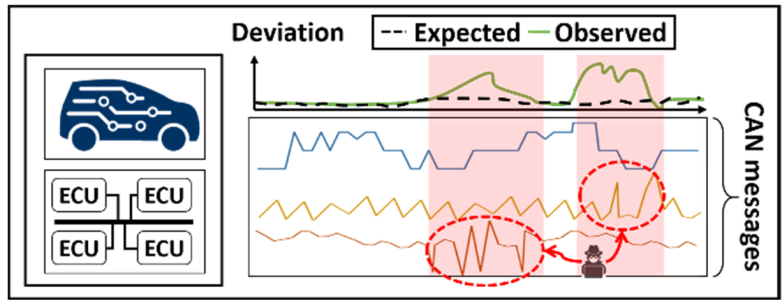

Figure 1: An example of an anomaly detection framework that monitors the network traffic and detects deviations from expected normal behavior during the attack intervals shown in red

## 2  RELATED WORK

Several automotive attacks have been studied by researchers to discover vulnerabilities in automotive systems. Recent attacks such as [15] exploit the vulnerability in security access algorithms to deploy airbags without any actual impact. The attackers in [16] reverse engineered a telematics control unit to exploit a memory vulnerability in the firmware to circumvent the existing firewall and remotely send diagnostic messages to control an ECU. Other attacks that compromised the ADAS camera sensor were studied in [17]. All of these attacks create anomalous behavior during vehicle operation, which a good anomaly detection framework must detect.

Anomaly detection has been a popular research topic in the domain of computer networks, and several solutions have been proposed to detect cyber-attacks in large-scale computer networks [18]. Although some of these solutions are highly successful in defending computer networks against various attacks, they require high compute power. The resource-constrained nature of automotive systems makes many of these solutions hard to adapt for detecting cyber-attacks in the in-vehicle networks. In the past decade, several solutions were developed to tackle the problem of anomaly detection in automotive systems [19]-[34]. These works can be broadly divided into two categories *(i)* heuristic-based, and *(ii)* machine learning based. Heuristic-based anomaly detection approaches typically observe for traces of known attack patterns, whereas a machine-learning-based approach can learn the normal behavior during an offline phase and observes for any deviation from the learned normal behavior at run-time, to detect anomalies. The heuristic-based techniques can be simple and have fast detection times when compared to machine learning based techniques. However, machine learning based techniques can detect both known and unknown attacks, which is not possible with heuristic based techniques. Some of the key prior works in these categories are discussed in the rest of this section.

## 2.1 Heuristic based anomaly detection

The authors in [19] used a language theory-based model to obtain signatures of known attacks from the vehicle's CAN bus. However, their approach fails to detect anomalous sequences when the model misses the packets transmitted during the early stages of an attack. In [20], the authors used transition matrices to detect anomalous sequences in a CAN bus-based system. This approach was able to achieve low false-positive rates for simple attacks but failed to detect realistic replay attacks. The authors in [21] proposed a Hamming-distance based model which monitors the CAN network to detect attacks. However, the model had very limited attack coverage. In [22], the authors proposed a specification-based approach and compared it with predefined attack patterns to detect anomalies. In [23], a time-frequency analysis model is used to continuously monitor CAN message frequency to detect anomalies. In [24], a heuristic-based approach is used to build a normal operating region by analyzing the messages at design time and using a message-frequency-based in-vehicle network monitoring system to detect anomalies at runtime. The authors in [25] use a clock-skew based fingerprint to detect anomalies by observing the variations in clock-skew of sender ECUs at runtime. In [26], the authors propose an anomaly detection system that monitors the entire system for change in entropy to detect anomalies. However, their approach fails to detect smaller anomalous sequences that result in minimal change in the entropy. *In a nutshell, heuristic-based anomaly detection systems provide low-cost and high-speed detection techniques but fail to detect complex and new attacks. Additionally, modeling every possible attack signature is practically impossible, and hence these anomaly detection approaches have a limited scope*.

## 2.2 Machine learning based anomaly detection

Recent works leverage advances in machine learning to build highly efficient anomaly detection systems. A deep neural network (DNN) based approach was introduced in [27], that continuously monitors the network and observes for change in communication patterns. However, this approach is only designed and tested for a low priority system (a tire pressure monitoring system), which limits us from directly adapting this technique to safety-critical systems. In [28], the authors proposed a recurrent neural network (RNN) based intrusion detection system that attempts to learn the normal behavior of CAN messages in the in-vehicle network. A hybrid approach was proposed in [29], which utilizes both specification and RNN based systems in two stages to detect anomalies. In [30] the authors propose an LSTM based predictor model that predicts the next time step message value at a bit level and detects intrusions by observing for large deviations in prediction errors. A long short-term memory (LSTM) based multi message-id detection model was proposed in [31]. However, the model is highly complex and has a high implementation overhead when deployed on an ECU. In [32], the authors proposed a GRU-based lightweight recurrent autoencoder and a static threshold-based detection scheme to detect various attacks in the in-vehicle network. The use of static threshold values for detection limits the scheme to detecting only very simple attacks. In [33], the authors propose a deep convolutional neural network (CNN) model to detect anomalies in the vehicle's CAN network. However, the model does not consider the temporal relationships between messages, which can better predict certain attacks. The authors in [34] proposed an LSTM framework with a hierarchical attention mechanism to reconstruct the input messages. A non-parametric kernel density estimator along with a k-nearest neighbors classifier is used to reconstruct the messages and the reconstruction error is used to detect anomalies. *Although most of these techniques attempt to increase the detection accuracy and attack coverage, none of them offers the ability to process very long sequences with relatively low memory and runtime overhead and still achieve reasonably high performance.*

In this paper, we propose a robust deep learning model that integrates a stacked LSTM based encoder-decoder model with a self-attention mechanism, to learn normal system behavior by learning to predict the next message instance. Table I summarizes some of the state-of-the-art anomaly detection works and their key features, and highlights the unique characteristics of our proposed LATTE framework. At runtime, we continuously monitor in-vehicle network messages and provide a reliable detection mechanism using a non-linear classifier. Sections 4 and 5 provide a detailed explanation of the proposed model and overall framework. In section 6 we show how our model is capable of efficiently identifying a variety of attack scenarios.

Table 1: Comparison between our proposed *LATTE* framework and the state-of-the-art works

| Technique | Task | Network architecture | Attention type | Detection technique | Requires labeled data? |
|---|---|---|---|---|---|
| BWMP [30] | Bit level prediction | LSTM network | - | Static threshold | Yes |
| RepNet [28] | Input recreation | Replicator network | - | Static threshold | No |
| HAbAD [34] | Input recreation | Autoencoder | Hierarchical | KDE and KNN | Yes |
| *LATTE* | Next message value prediction | Encoder-decoder | Self-attention | OCSVM | No |

## 3 BACKGROUND

Solving complex problems using deep learning was made possible due to advances in computing hardware and the availability of high-quality datasets. Anomaly detection is one such problem that can leverage the power of deep learning. In an automotive system, ECUs exchange safety-critical messages periodically over the in-vehicle network. This time series exchange of data results in temporal relationships between messages, which can be exploited to detect anomalies. However, this requires a special type of neural network, called Recurrent Neural Network (RNN) to capture the temporal dependencies between messages. Unlike traditional feed-forward neural networks where the output is independent of any previous inputs, RNNs use previous sequence state information in computing the output, which makes them an ideal choice to handle time-series data.

### 3.1 Recurrent Neural Network (RNN)

An RNN [35] is the most basic sequence model that takes sequential data such as time-series data as the input and learns the underlying temporal relationships between data samples. An RNN block consists of an input, an output, and a hidden state that allows it to remember the learned temporal information. The input, output, and hidden state all correspond to a particular time step in the sequence. The hidden-state information can be thought of as a data point in the latent space that contains important temporal information about the inputs from previous time steps. The current stage output of an RNN is computed by taking the previous hidden-state information along with the current input. Moreover, since the backpropagation in RNNs occurs through time, the error value shrinks or grows rapidly leading to vanishing or exploding gradients. This severely restricts RNNs from learning patterns in the input data that have long-term dependencies [36]. To overcome this problem, long short-term memory (LSTM) networks [37] with additional gates and states were introduced.

### 3.2 Long Short-Term Memory (LSTM) network

LSTMs are enhanced RNNs that consist of a cell state, hidden state, and multiple additional gates that help in learning long-term dependencies. The cell state carries the relevant long-term dependencies throughout the processing of an input sequence, whereas the hidden state contains relevant information from the recent time steps accommodating short-term dependencies. The gates in LSTM regulate the flow of the information from the hidden state to the cell state. These combinations of gates and states give LSTM an edge over the simple RNN in remembering long-term dependencies in sequences. LSTMs have therefore replaced simple RNNs in the areas of natural language processing, time-series forecasting, and machine translation [36].

In general, LSTMs overcome many of the limitations of RNNs and provide a more than acceptable solution for the vanishing and exploding gradient descent problems. However, their performance drops significantly when handling very long sequences (e.g., with 100 or more time steps). This is mainly because the predictions of an LSTM unit at the current time step $t$, are heavily influenced by its previous hidden state and cell state at time step $t-1$ as compared to the past time steps. Therefore, for a very long input sequence, the representation of the input at the first time step tends to diminish as the LSTM processes inputs at the future time steps. To overcome this limitation, we need a mechanism that can look back and identify the information that can influence future sequences. One such look-back mechanism is neural attention, which is discussed next.

### 3.3 Attention

Attention, or neural attention is a mechanism in neural networks that can mimic the visual attention mechanism in humans [38]. A human eye can focus on certain objects or regions with higher resolution compared to their surroundings. Similarly, the attention mechanism in neural networks can allow focusing on the relevant parts of the input sequence and selectively output only the most relevant information. While sequence models such as LSTMs typically take the previous hidden state information and the input at the current time step to compute the current output, they suffer in performance when processing very long input sequences as the information from the first time step is less representative in the hidden states compared to the information from the very recent time steps. Incorporating attention mechanisms with LSTMs can overcome this problem by allowing the sequence models to capture the crucial information from any past time steps of the input sequence.

Attention mechanisms are frequently used in encoder-decoder architectures [36]. An encoder-decoder architecture mainly consists of three major components *(i)* encoder, *(ii)* latent vector, and *(iii)* decoder. The encoder converts the input sequence to a fixed-size latent representation called a latent vector. The latent vector contains all the information representing the input sequence in a latent space. The decoder takes the latent vector as input and converts it to the desired output. However, due to the latent vector's fixed-length representation of the input sequence, it fails to encapsulate all the information from a very long input sequence, thereby resulting in poor performance. To address this problem, the authors in [39] introduced an attention mechanism in sequence models that enabled encoders to build a context vector by creating customized shortcuts to parts of the inputs. This ensures that the context vector represents the crucial parts and learns the very long-term dependencies in the input sequence leading to improved decoder outputs. In [40], the authors propose a self-attention mechanism for an LSTM encoder-decoder model that consumes all the encoder hidden states to compute the attention weights.

An illustration of generating the input to the decoder in an LSTM-based encoder-decoder model rolled out in time for 4 time steps is shown in Fig. 2. The input to the LSTM at each time step is represented as ($x_t$) and the initial hidden vector is $h_0$. The colored rectangle next to each LSTM unit for every time step represents the hidden state information and the height of each color signifies the amount of information from each time step. Inside the LSTM cell at each time step, a square filled with a different color is used to represent the hidden state information of that time step. Moreover, for this example, we consider a scenario where the output at the last time step ($t=4$) has a high dependency on the input at the second time step ($x_2$). We can see that in Fig. 2(a), the LSTM hidden state at $t=4$ largely comprises of information from the third (blue) and fourth (orange) time steps. This results in sending the decoder an incorrect representation of current time step dependency, which leads to poor results at the output of the decoder. On the contrary, in Fig. 2(b), the self-attention block consumes all hidden state representations at each time step as well as the current time step ($t=4$) and generates the context vector (decoder input). It can be observed that the self-attention mechanism clearly captures the high dependency of output at $t=4$ on the output at $t=2$ (shown in the hidden state information at the output of self-attention). This can also be seen in the attention weights computed by the self-attention where the information from the second (green) time step is given high weightage compared to others. Therefore, by better representing the important parts of the input sequence in the decoder input, the self-attention mechanism is able to facilitate better decoder outputs. Also, unlike other attention mechanisms such as [41], the attention vector in self-attention aligns encoder outputs to encoder hidden states, thereby removing the need for any feedback from previous decoder predictions. Moreover, due to the lack of a decoder feedback loop, the self-attention mechanism can quickly learn the temporal dependencies in the long input sequences. In this work, for the first time, we adapt the self-attention mechanism to a stacked LSTM based encoder-decoder network to learn the temporal relationships between messages in a CAN based automotive network.

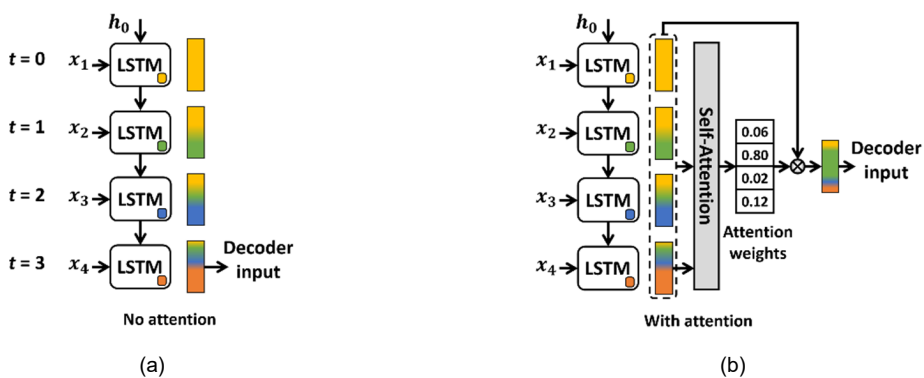

Figure 2: Comparison of input to the decoder in case of (a) no attention, (b) with attention in sequence models using LSTMs

## 4 PROBLEM FORMULATION

### 4.1 System Overview

In this work, we consider an automotive system that consists of multiple ECUs connected using a CAN based in-vehicle network, as shown in Fig. 3. Each ECU consists of three major components: *(a)* processor, *(b)* communication controller, and *(c)* transceiver. A processor can have single or multiple cores that are used to execute real-time automotive applications. Most of these automotive applications are hard real-time and have strict timing and deadline constraints. Each application can be modeled as a set of data dependent and independent tasks mapped to different ECUs. The dependent tasks communicate by exchanging messages over the CAN network. A communication controller acts as the interface between the computation and communication realms. It facilitates the data movement from the processor to the network fabric and vice versa. Some of the important functions of a communication controller include packing of data from the processor into CAN frames, managing the transmission and reception of CAN frames, and filtering CAN messages based on the pre-programmed CAN filters (done by the original equipment manufacturer (OEM) when programming the communication controller). Lastly, a transceiver acts as an interface between the physical CAN network and the ECU, and facilitates the transmission and reception of CAN frames to and from the network respectively. In this work, we do not consider monitoring the execution within the CAN hardware IPs as it would require access to proprietary information that is only available to OEMs. We therefore assume that the proprietary CAN hardware IPs are "black boxes" and design an anomaly detection solution that does not require the complexity that comes with monitoring the internals of these IP blocks. This assumption is also consistent with all prior works on in-vehicle network anomaly detection work. However, if we were able to get access to this hardware stack and the program execution on the CAN hardware IP, our framework can be extended to analyze CAN IPs and detect the attacks before they appear on the in-vehicle network.

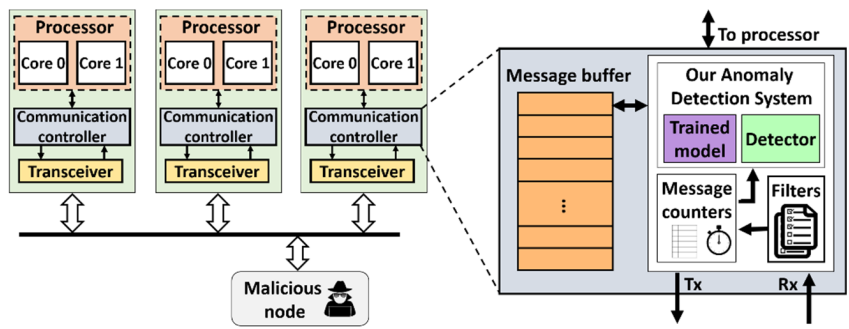

Figure 3: Overview of the system model with our proposed modifications to the communication controller

To accommodate anomaly detection, we require modifications to existing CAN communication controllers. A traditional CAN communication controller consists of message filters that are used to filter out unwanted CAN messages and message buffers to temporarily store the messages before they are sent to the processor. This can be observed in the right region of Fig. 3. We introduce message counters to this controller, which take the output of the message filters and keep a track of message frequencies. This bookkeeping helps in the observation of any abnormal message rates that may occur during a distributed denial of service (DDoS) attack (see section 4.3). After confirming the message rate, the message is sent to the deployed anomaly detection system where it goes through a two-step process to determine whether the message is anomalous or not.

In the first step, our trained LSTM based attention model is used to predict the next message instance, which is then used to compute the deviation from the true message. This deviation measure is given as the input to a detector unit that uses a non-linear classifier to determine if a given deviation measure represents a normal or an anomalous message. The details related to the models and the deviation metrics used in our framework are discussed in detail in sections 5.2 and 5.3 respectively. Messages are temporarily stored in the message buffer before they are validated and sent to the processor. If the anomaly detection system determines a particular message to be anomalous, it is discarded from the buffer and will not be sent to the processor, thereby avoiding the execution of attacker messages.

*Note:* Our anomaly detection system is implemented in the communication controller instead of a centralized ECU to *(i)* avoid single-point failures, *(ii)* prevent scenarios where the in-vehicle network load increases significantly due to high message injection (e.g., due to a DDoS attack, explained in section 4.3), where the centralized ECU will not be able to communicate with a target ECU, and *(iii)* enable independent and immediate detection without delay compared to relying on a message from a centralized ECU. Lastly, we chose the communication controller instead of the processor to avoid jitter in real-time application execution.

### 4.2 Communication Overview

In this work, we consider Controller Area Network (CAN) as the in-vehicle network protocol that is used for exchanging time-critical messages between ECUs. CAN is a lightweight, low-cost, event-triggered in-vehicle network protocol, and is the defacto industry standard. Several variants of CAN have been proposed over time, but the CAN standard 2.0B remains the most popular and widely used in-vehicle network protocol till today.

A CAN message consists of one or multiple signal values. Each signal contains independent information corresponding to a sensor value, actuator control, or computation output of a task on an ECU. Signals are grouped with additional information to form CAN frames. Each CAN frame mainly consist of a header, payload, and trailer segments (Fig. 4). The header consists of an 11-bit (CAN standard) or 29-bit (CAN extended) unique message identifier and a 6-bit control field. This is followed by a 64-bit payload segment and a 15-bit cyclic redundancy check (CRC) field in the trailer segment. The payload segment consists of multiple signals that are arranged in a predetermined order as per the definitions in the CAN database (.dbc) files. In addition, the CAN frame also has a 1-bit start of the frame (SOF) field at the beginning of the header, two 1-bit delimiters separating the 1-bit acknowledgment (ACK) field, and a 7-bit end of frame (EOF) field in the trailer segment.

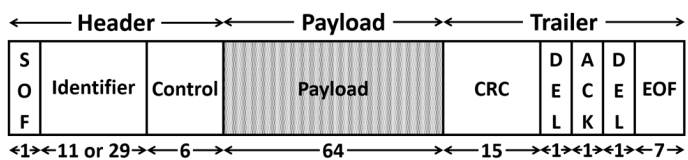

Figure 4: Controller Area Network (CAN) 2.0B communication frame

In this work, our proposed anomaly detection framework operates on the payload segment of the CAN frame i.e., signals within each message. The main motivation for monitoring the payload field is because the attacker needs to modify the bits in the payload to launch any attack (a modification in the header or trailer segments would simply result in the frame getting invalidated at the receiving ECU). Our proposed *LATTE* framework learns the temporal dependencies between the message instances at design time by learning to predict the next message instances and observe for deviations at runtime to detect cyber-attacks. Moreover, as our framework mainly focuses on monitoring of the payload field, our technique is agnostic to the in-vehicle network protocol and can be extended to other in-vehicle network protocols such as CAN-FD, FlexRay, etc., with minimal changes. The details related to the detection of cyber-attacks using our proposed anomaly detection system are presented in sections 5.2 and 5.3.

### 4.3 Threat Model

We assume that the attacker can gain access to the in-vehicle network using the most common threat vectors such as connecting to the vehicle OBD-II port, probing into the in-vehicle network, and via advanced threat vectors such as connected V2X ADAS systems, insecure infotainment systems, or by replacing a trusted ECU with a malicious ECU. We also assume that the attacker has access to the in-vehicle network parameters such as flow control, BAUD rate, parity, channel information, etc. that can be obtained by a simple CAN data logger, and can help in the transmission of malicious messages. We further assume a pessimistic situation where the attacker can access the in-vehicle network at any instance and try to send malicious messages.

Given the above assumptions, our proposed anomaly detection system tries to protect the in-vehicle network from the multiple types of cyber-attacks listed below. These attacks are modeled based on the most common and hard-to-detect attacks in the automotive domain.

1. *Constant attack*: In this attack, the attacker overwrites the signal value to a constant value for the entire duration of the attack interval. The complexity of detection of this attack depends on the change in magnitude of signal value. Intuitively, a small change in the magnitude of the signal value is harder to detect than larger changes.

2. *Continuous attack*: In this attack, the attacker tries to trick the anomaly detection system by continuously overwriting the signal value in small increments until a target value is achieved. The complexity of detecting this attack depends on the rate of change of the signal value. Larger change rates are easier to detect than smaller rates.

3. *Replay attack*: In this attack, the attacker plays back a valid message transmission from the past, tricking the anomaly detector into believing it to be a valid message. The complexity for detecting this attack depends mainly on the frequency and sometimes on the duration of the playbacks. High-frequency replays are easier to detect compared to low-frequency replays.

4. *Dropping attack*: In this attack, the attacker disables the transmission of a message or group of messages resulting in missing or dropping of communication frames. The complexity of detecting this attack depends on the duration for which the messages are disabled. Longer durations are easier to detect due to missing message frames for a prolonged time compared to shorter durations.

5. *Distributed Denial of Service (DDoS) attack*: In this attack, the attacker floods the in-vehicle network with an arbitrary or specific message with the goal of increasing the overall bus load and rendering the bus unusable for other ECUs. This is the most common and easy to launch attack as it requires no information about the nature of the message. These attacks are fairly simple to detect even using a rule-based approach as the message frequencies are fixed and known at design time for automotive systems. Any deviation in this message rate can be used as an indicator for detecting this attack.

*Problem objective*: The main objective of our work is to develop a real-time anomaly detection framework that can detect various cyber-attacks in CAN-based automotive networks, that has *(i)* high detection accuracy, *(ii)* low false-positive rate, *(iii)* high precision and recall, *(iv)* large attack coverage, and *(v)* minimal implementation overhead (low memory footprint, fast runtime) for practical anomaly detection in resource-constrained ECUs.

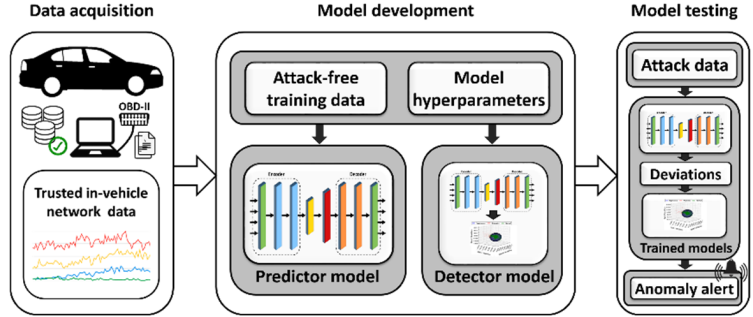

Figure 5: Overview of proposed *LATTE* framework

## 5 PROPOSED FRAMEWORK

An overview of our proposed *LATTE* framework is shown in Fig. 5. Our framework consists of a novel self-attention based LSTM deep learning model that is trained with data obtained from a data acquisition step. The data acquisition step collects trusted in-vehicle network data under a controlled environment. We then post-process and use this data to train the stacked LSTM self-attention predictor model in an unsupervised setting to learn the normal operating behavior of the system. We also developed a one class support vector machine (OCSVM) based detector model that utilizes the predictions from the LSTM predictor to detect cyber-attacks as anomalies at run-time. After training, the framework is tested by being subjected to various attacks. The details of this framework are presented in the subsequent subsections.

### 5.1 Data acquisition

This is the first step of the *LATTE* framework and involves collecting the in-vehicle network data from a trusted vehicle. It is important to ensure that the in-vehicle network and the ECUs in the vehicle are free from the attackers. This is because the presence of an attacker can result in logging corrupt in-vehicle network data that falsely represents the normal operating conditions, leading to learning an inaccurate representation of the normal system behavior with our proposed models. Moreover, it is also crucial to cover a wide range of normal operating conditions and have the data collected over multiple intervals, to ensure high confidence in the collected data. The performance of the anomaly detection system is highly dependent on the quality of the collected data, and thus this is a crucial step. Additionally, the type of data collected depends on the functionalities or ECUs that are subjected to monitoring by the anomaly detection system. The most common access point to collect the in-vehicle network data is the OBD-II port, which gives access to the diagnostic and most commonly used messages. However, we recommend probing into the CAN network and logging the messages, as it gives unrestricted access to the in-vehicle network, unlike the OBD-II port.

After collecting the message data from the in-vehicle network, the data is prepared for pre-processing to make it easier for the training models to learn the temporal relationships between messages. The full dataset is split into groups based on the unique CAN message identifier and each group is processed independently. The data entries in the dataset are arranged as rows and columns with each row representing a single data sample corresponding to a particular timestamp and each column representing a unique feature of the message. The columns consist of the following features: *(i)* timestamp at which the message was logged, *(ii)* message identifier, *(iii)* number of signals in the message, *(iv)* individual signal values (one per column), and *(v)* a single bit representing the label of the message. The label column is 0 for non-anomalous samples and 1 for anomalous samples. The label column is set to 0 for all samples in the training and validation dataset as all the data samples are non-anomalous and collected in a trusted environment. The label column will have a value of 1 for the samples in the test dataset during the attack interval and 0 for the other cases. However, it is important to highlight that we do not use this label information while training our predictor and detector models. Moreover, for each signal type, the signal values are scaled between 0 to 1 as there can be a high variance in the signal magnitudes. Such high variance in the input data can result in very slow or unstable training. Additionally, in this work, we do not consider timestamp as a unique feature. We use the concept of time in a relative manner when training (to learn patterns in sequences) and during deployment. We are not dependent on absolute time during training and deployment. We use the dataset presented in [31] to train and evaluate our proposed LATTE framework. The dataset consists of both normal and attack CAN message data. Details related to the models and the training procedure are discussed in the next subsections, while the dataset is discussed in section 6.1.

### 5.2 Predictor model

We designed predictor and detector models that work in tandem to detect cyber-attacks as anomalies in the in-vehicle network. The predictor model attempts to learn the normal system behavior via an unsupervised learning approach to predict the next message instance with high accuracy at design time using the normal (non-anomalous) data. During this process, the predictor model learns the underlying distribution of the normal data and relates it to the normal system behavior. This knowledge of the learned distribution is used to make accurate predictions of the next message instances at runtime for normal messages. In the event of a cyber-attack, the message values no longer represent the learned distribution or maintain the same temporal relationships between messages, leading to large deviations between the predictions and the true (observed) messages. In this work, we employ a non-linear classifier based detector model to learn the deviation patterns that correspond to the normal messages, which is then used to detect anomalies (i.e., attacks that cause anomalous deviations) at runtime. The details related to the detector model are discussed in detail in section 5.3.

Our proposed predictor model consists of a stacked LSTM based encoder-decoder architecture with the self-attention mechanism. This is illustrated in Fig. 6. The first linear layer in the predictor model takes the time series CAN message data as the input and generates a 128 dimensional embedding for each input. Each input sample consists of *k* features where each feature represents a particular signal value within that message. The output embedding from the linear layer is passed to the stacked two-layer LSTM encoder to produce a 64-dimension encoder output ($h_1^e, h_2^e ... h_t^e$). The encoder output is the latent representation of the input time-series signal values that encompass the temporal relationships between messages. The self-attention block generates the context vector ($\varphi_t$) by applying the self-attention mechanism to the encoder outputs. The self-attention mechanism begins by applying a linear transformation on the encoder's current hidden state ($h_t^e$) and multiplies the result with the encoder output. The output from the multiplication is passed through a softmax activation to compute the attention weights. The attention weights represent the importance of each hidden state information from the earlier time steps, at the current time step. The attention weights are scalars multiplied with the encoder outputs to compute the attention applied vector ($a_n$) which is then combined with the encoder output to compute the input to the decoder (context vector ($\varphi_t$)). The context vector along with the previous decoder's hidden state ($h_{t-1}^d$) is given as input to the stacked two-layer decoder, which produces a 64-dimension output that is passed to the last linear layer to obtain a *k* dimensional output. This *k* dimension output represents the signal values of the next message instance. Thus, given an input sequence $X = \{x_1, x_2, \ldots x_t\}$, our predictor model predicts the sequence $\hat{X} = \{\hat{x}_1, \hat{x}_2, \ldots, \hat{x}_t\}$, where the output at time step *t* ($\hat{x}_t$) is the prediction of the input at time step *t+1* ($x_{t+1}$). The last prediction ($\hat{x}_t$) is generated by consuming the complete input sequence (*X*).

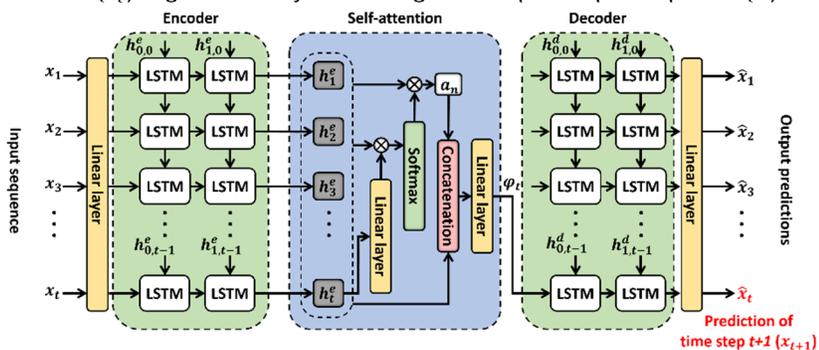

Figure 6: Our proposed predictor model for the *LATTE* anomaly detection framework showing the stacked LSTM encoder – decoder rolled out in time for *t* time steps along with the self-attention mechanism generating context vector for time step *t*. The output at time step *t* ($\hat{x}_t$) is the prediction of the input at time step *t+1* ($x_{t+1}$).

This predictor model is trained using non-anomalous (normal) data without any labels in an unsupervised manner. To train the model with sequences, we employ a rolling window approach. We consider a window of fixed size length (known as subsequence length) consisting of signal values over time. The window with signal values is called a subsequence and has subsequence length number of samples of signal values. Our predictor model learns the temporal dependencies that exist between the signal values within the subsequence and uses them to predict the signal values in the next subsequence (i.e., window shifted to the right by one-time step). The signal values corresponding to the last time step in the output subsequence represent the final prediction, as the model consumes the entire input subsequence to generate them. We compare this last time step in the output subsequence with the actual signal values and compute the prediction error using the mean square error (MSE) loss function. This process is repeated until the end of the training dataset. The subsequence length is a hyperparameter related to the LSTM network and is independent of the vehicle and the message data, and need to be selected before training the model. We conducted multiple experiments with different model parameters, and selected the hyperparameters that gave us the best performance results. The predictor model is trained by splitting the dataset into training (80%) and validation (20%) data without shuffling, as shuffling would destroy the existing temporal relationships between messages. During the training process, the model tries to minimize the prediction error in each iteration (a forward and backward pass) by adjusting the weights of the neurons in each layer using backpropagation through time. At the end of each training epoch, the model is validated (forward pass only) using the validation dataset to evaluate the model performance. We employ mini-batches to speed up the training process and use an early stopping mechanism to avoid overfitting. The details related to the non-anomalous dataset and the hyperparameters selected for the model are presented in section 6.1.

### 5.3 Detector model

After training the predictor model, we train a separate classifier (detector model) that utilizes the information from the predictor to detects attacks. The anomaly detection problem can be treated as a binary classification problem as we are mainly interested in distinguishing between normal and anomalous messages. In general, as the in-vehicle network data recordings can grow in size very rapidly, labeling this data can get very expensive. Additionally, due to the nature of the frequency of attack scenarios, the number of attack samples would be quite small compared to normal samples even when the dataset is labeled. This results in having a highly imbalanced dataset that would result in poor performance when trained with a traditional binary classifier in a supervised learning setting. However, a popular non-linear classifier known as a support vector machine (SVM) can be altered to make it work with unbalanced datasets where there is only one class. Hence, in this work, we use a one class support vector machine (OCSVM) to classify the messages as anomalous or normal. The OCSVM learns the distribution of the training dataset by constructing the smallest hypersphere that contains the training data at design time and identifies any sample outside the hypersphere as an anomaly at runtime. We train an OCSVM by using the output from the previously trained predictor model. We begin by giving the previously used normal training dataset as the input to the predictor to generate the predictions. We then compute the deviations (prediction errors) for all the training data and pass it as input to the OCSVM. The OCSVM tries to generate the smallest hypersphere that can fit most of the deviation points and uses it at runtime to detect anomalies. Fig. 7 shows an example of a hypersphere generated by training an OCSVM for a message with three signals. Each axis in the figure represents the relevant signal deviation and the dark blue sphere represents the decision boundary. It can be observed that almost the entirety of training data (shown via green dots) is confined to within the blue sphere.

In our work, the deviation of a message is represented as a vector where each element of the vector corresponds to the difference between the true and predicted signal value. Therefore, for a message *m* with $k_m$ number of signals, the deviation vector ($\Delta_{m,t}$) computed at time step *t* is given by equation (1).

$$\Delta_{m,t} = (\hat{S}_{i,t} - S_{i,t+1}) \in \mathbb{R}^2, \forall i \in [1, k_m] \quad (1)$$

where $\hat{S}_{i,t}$ represent the prediction of the next true $i^{th}$ signal value ($S_{i,t+1}$) made at time step *t*. We also experimented with other deviation measures that are given by equations (2), (3) and (4).

$$\Delta_{m,t}^{sum} = \sum_{i=1}^{k_m} |\Delta_{m,t}|, \forall i \in [1, k_m] \quad (2)$$

$$\Delta_{m,t}^{avg} = \frac{1}{k_m} \sum_{i=1}^{k_m} |\Delta_{m,t}|, \forall i \in [1, k_m] \quad (3)$$

$$\Delta_{m,t}^{max} = \max(|\Delta_{m,t}|), \forall i \in [1, k_m] \quad (4)$$

Moreover, there can be situations where some of the signal deviations in a message can be positive while others are negative. This could potentially result in making the sum or mean of signal deviations zero or near zero, falsely representing no deviation or very small deviation. To avoid these situations, we use the absolute signal deviations to compute the deviations for the variants. *Note*: Unlike equation (1) that uses a vector of *k* dimensions to represent the message deviation, equations (2), (3), and (4) reduce the vector to a single value using different reduction operations. We explored these reduced deviation scores (shown in equations (2), (3), and (4)) that utilize absolute deviation values to determine the best one, as discussed in section 6.2.

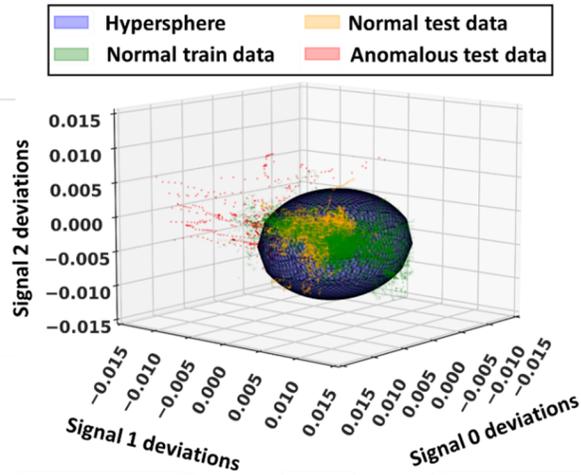

Figure 7: OCSVM decision boundary shown in the blue sphere with the green dots showing the normal samples from training data, and yellow and red dots showing the normal and anomalous samples respectively from test data

In summary, our predictor model predicts the normal samples with very small deviations and anomalous samples with high deviations. The OCSVM takes this predictor property into account when constructing the hypersphere. In Fig. 7, the yellow dots and red dots represent the normal and anomalous samples respectively in the test dataset. It can be observed that when the test data with anomalies is given as input to the OCSVM, it generally correctly classifies the yellow samples within the hypersphere and red samples outside the hypersphere. Thus, both predictor and detector models work collectively to detect attacks as anomalies. The details related to the testing process are described in the next subsection.

### 5.4 Model testing

In the deployment/testing step, we present a test dataset consisting of anomalous samples representing multiple attacks (outlined in section 4.3) along with the normal samples to the *LATTE* framework. The normal messages have a label value of 0 and the attack messages have a label value of 1. During this step, each sample (signal values in a message) is first sent to the predictor model to predict the signal values of the next message instance, and the deviation is computed based on the true message data. This deviation vector is passed to the OCSVM detector model, to compute the position of the deviation vector in the *k*-dimensional space, where *k* represents the number of signals in the message. The message is marked as non-anomalous when the point corresponding to the deviation vector falls completely inside the learned hypersphere. Otherwise, the message is marked as anomalous and an anomaly alert is raised. This can be used to invoke an appropriate remedial action to suppress further actions from the attacker. However, the design of remedial actions and response mechanisms falls outside the scope of our paper. The performance evaluation of our proposed *LATTE* framework under various attack scenarios is presented in detail in sections 6.2 and 6.3.

## 5.5 Anomaly detection system deployment

Our proposed anomaly detection system can be deployed in a real-world vehicle in two different approaches. The first is a global monitoring or centralized approach, where a powerful centralized ECU monitors the messages on the CAN bus and detects anomalies. The second approach involves distributing the anomaly detection task to across ECUs and only monitoring the messages that are relevant to that particular ECU (distributed monitoring). Both choices have pros and cons, but we believe that the distributed monitoring has multiple advantages over the centralized approach because of the following reasons.

- A centralized approach is prone to single-point failures, which can completely open up the system to the attacker;
- In extreme scenarios such as during a DDoS attack (explained in section 4.3), the in-vehicle network can get highly congested and the centralized system might not be able to communicate with the victim ECUs;
- If an attacker succeeds in fooling the centralized ECU, attacks can go undetected by the other ECUs, resulting in compromising the entire system; whereas with a distributed detection scheme, fooling multiple ECUs is required which is much harder, and even if an ECU is compromised, this can still be detected by the decentralized intelligence in a distributed detection;
- In a distributed detection, ECUs can stop accepting messages as soon as an anomaly is detected without waiting for a centralized system to notify them, leading to faster response;
- The computation load of detection is split among the ECUs with a distributed approach, and the monitoring can be limited to only the required messages. Thus, multiple ECUs can monitor a subset of messages independently, with lower overhead;

Many prior works, e.g., in [19] and [24], consider a distributed local detection approach for these reasons. Moreover, with automotive ECUs becoming increasingly powerful, the collocation of detection tasks with real-time automotive applications in a distributed manner should not be a problem, provided the overhead from the detection is minimal. The light weight nature and anomaly detection performance of our proposed *LATTE* framework are discussed further in section 6. Moreover, as the detector looks at the payload segment individually, it needs to keeps a track of the previous messages to detect anomalous patterns. We can cache the previous normal samples and predictions (in the case of anomalies) and use them to preserve the dependencies within the data, which can be later used in determining whether the next sample is normal or anomalous. To minimize the storage overhead, we can employ a circular buffer of size equal to the subsequence length (configured at design time). Using this approach, we can still look into the message dependencies in the past.

## 6 EXPERIMENTS

### 6.1 Experimental Setup

To evaluate the effectiveness of our proposed *LATTE* framework, we first explored five variants of the same framework with different deviation criteria: *LATTE*-ST, *LATTE*-Diff, *LATTE*-Sum, *LATTE*-Avg, and *LATTE*-Max. *LATTE*-ST uses our proposed predictor model with a static threshold (ST) value to determine whether a given message is anomalous or normal based on the deviation. The other four variants use the same predictor model but different detection criteria for computing the deviations for OCSVM. *LATTE*-Diff uses the difference in signal values (equation (1)); *LATTE*-Sum and *LATTE*-Avg use a sum and mean of absolute signal deviations respectively (equations (2), and (3)); and *LATTE*-Max uses the maximum absolute signal deviation (equation (4)), as the input to the detector model.

Subsequently, we compare the best variant of our framework with four prior works: Bitwise Message Predictor (BWMP [30]), Hierarchical Attention-based Anomaly Detection (HAbAD [34]), a variant of [34] called Stacked HAbAD (S-HAbAD [34]), and RepNet [28]. BWMP [30] trains an LSTM based neural network that aims to predict the next 64 bits of a CAN message by minimizing the bitwise prediction error using a binary cross-entropy loss function. At runtime, BWMP uses the prediction loss as a measure to detect anomalies. HAbAD [34] uses an LSTM based autoencoder model with hierarchical attention. The HAbAD model attempts to recreate the input message sequences at the output and aims to minimize reconstruction loss. Additionally, HAbAD uses supervised learning in the second step to model a detector using the combination of a non-parametric kernel density estimator (KDE) and k-nearest neighbors (KNN) algorithm to detect cyber-attacks at runtime. Lastly, S-HAbAD is a variant of HAbAD that uses stacked LSTMs as autoencoders and uses the same detection logic used by the HAbAD. The S-HAbAD variant is compared against to show the effectiveness of using stacked LSTM layers. Lastly, RepNet [28] uses simple RNNs to increase the dimensionality of input signal values and attempts to reconstruct the signal values at the output by minimizing the reconstruction error using mean squared error. At runtime, RepNet monitors for large reconstruction errors to detect anomalies. The results of all experiments are discussed in detail in subsections 6.2-6.4.

We conducted all experiments using an open-source CAN message dataset developed by ETAS and Robert Bosch GmbH [31]. The dataset consists of CAN message data for different message IDs consisting of various fields such as timestamps, message ID, and individual signal values. Additionally, the dataset consists of a training dataset with only normal data and a labeled test dataset with multiple attacks (as discussed in section 4.3). The attack data in the dataset is modeled from the real world attacks that are commonly seen in automotive systems. It is important to note that we do not use any labeled data during the training or validation of our models and learn the normal system behavior in an unsupervised manner. The labeled data is given to the models only during the testing phase and used to compute performance metrics. Moreover, the dataset consists of multiple message frequencies {15, 30, 45} ms. Since the high frequency messages pose a significant challenge to the anomaly detection system, and could result in high overhead, in this work we consider the message frequency of 15 ms for all of our experiments.

We used PyTorch 1.5 to implement all of the machine learning models including *LATTE* and its variants, and the models from the comparison works. Our proposed predictor model is trained with 80% of the available normal data and the remaining 20% is used for validation. We conducted multiple experiments with different model parameters, and selected the hyperparameters that gave us the best performance results. The training phase is repeated for 500 epochs with an early stopping mechanism that monitors the validation loss after the end of each epoch and stops if there is no improvement after 10 (patience) epochs. We used the ADAM optimizer with mean squared error (MSE) as the loss function. Additionally, we employed a rolling window approach (discussed in section 5.2) with a subsequence length of 32 time steps, a batch size of 256, and a starting learning rate of 0.0001. We used the scikit-learn package to implement the OCSVM in the detector model (section 5.3). We used a radial basis function (RBF) kernel with a kernel coefficient (*gamma*) equal to the reciprocal of the number of features (i.e., number of signals in the message). Moreover, to speed up OCSVM training, we set the kernel cache size to 400 MB and enabled the shrinking technique to avoid solving redundant optimizations. All the simulations are run on an AMD Ryzen 9 3900X server with an Nvidia GeForce RTX 2080Ti GPU.

Before looking at the experimental results for various performance metrics, it is important to understand some key definitions in the context of anomaly detection. We define a true positive as the scenario when an actual attack is detected as an anomaly by the anomaly detection system and a true negative as the situation where an actual normal message is detected as normal. Additionally, a false positive would be a false alarm where a normal message is incorrectly classified as an anomaly and a false negative would occur when an anomalous message is incorrectly classified as normal. Using the above definitions, we evaluate our proposed framework using four different metrics: *(i) Detection accuracy*: a measure of the anomaly detection system's ability to detect anomalies correctly, *(ii) False positive rate*: i.e., false alarm rate, *(iii) F1 score*: a harmonic mean of precision and recall; we use the F1-score instead of individual precision and recall values as it captures the combined effect of both precision and recall metrics, and *(iv) receiver operating characteristic (ROC) curve with area under the curve (AUC)*: a popular measure of classifier performance. A highly efficient anomaly detection system has high detection accuracy, F1 score, and ROC-AUC while having a very low false-positive rate.

## 6.2 Comparison of *LATTE* variants

In this subsection, we present the comparison results of the five variants *LATTE*-ST, *LATTE*-Sum, *LATTE*-Avg, *LATTE*-Max and *LATTE*-Diff. All the variants of *LATTE* use the trained predictor model (discussed in section 5.2) to make the predictions and use OCSVM as a detector except in the case of *LATTE*-ST, which uses a fixed threshold scheme introduced in [32] to predict the given message as normal or anomalous. The main purpose of this experiment is to analyze the impact of using a non-linear classifier such as OCSVM on the model performance instead of a simple static threshold scheme (*LATTE*-ST). Additionally, with the last four variants, we aim to study the effect of different deviation criteria on the OCSVM detection performance. The deviations for any given message in *LATTE*-Diff ($\Delta_{m,t}$), *LATTE*-Sum ($\Delta_{m,t}^{sum}$), *LATTE*-Avg ($\Delta_{m,t}^{avg}$) and *LATTE*-Max ($\Delta_{m,t}^{max}$) are computed using the equations (1), (2), (3) and (4) respectively.

Fig. 8(a)-(c) shows the detection accuracy, false-positive rate, and F1 score respectively for the five different variants of *LATTE* under five different attack scenarios discussed in section 4.3. The 'No attack' case involves testing the model with new non-anomalous data that the model has not seen before. Firstly, from Fig. 8(a)-(c) it is clear that the OCSVM based detection models clearly outperform the static threshold models (*LATTE*-ST). This is mainly because of their ability to process complex attack patterns and generate non-linear decision boundaries that can distinguish better between normal and anomalous data. Moreover, it can be seen that *LATTE*-Diff outperforms all the OCSVM based models in detection accuracy, false-positive rate, and F1 score. Lastly, in Fig. 8(d), we present the ROC curves and the corresponding AUC values in the brackets next to each legend. Out of the various attacks, we show results for continuous attacks, as it is the most challenging attack to detect. This is because during this attack, the attacker constantly tries to fool the anomaly detection system into thinking that the signal values in the messages are legitimate. This requires careful monitoring and the ability to learn complex patterns to differentiate between normal and anomalous samples. On average, across all attacks, LATTE-Diff was able to achieve an average of 13.36% improvement in accuracy, 11.34% improvement in F1 score, 17.86 % improvement in AUC and 47.9% reduction in false positive rate, and up to 42% improvement in accuracy, 32.6% improvement in F1 score, 29.4% improvement in AUC and 95% decrement in false positive rate, compared to the other variants. Therefore, we selected *LATTE*-Diff as our candidate model for subsequent experiments where we present comparisons with the state-of-the-art anomaly detectors. Henceforth, we refer to *LATTE*-Diff as *LATTE*.

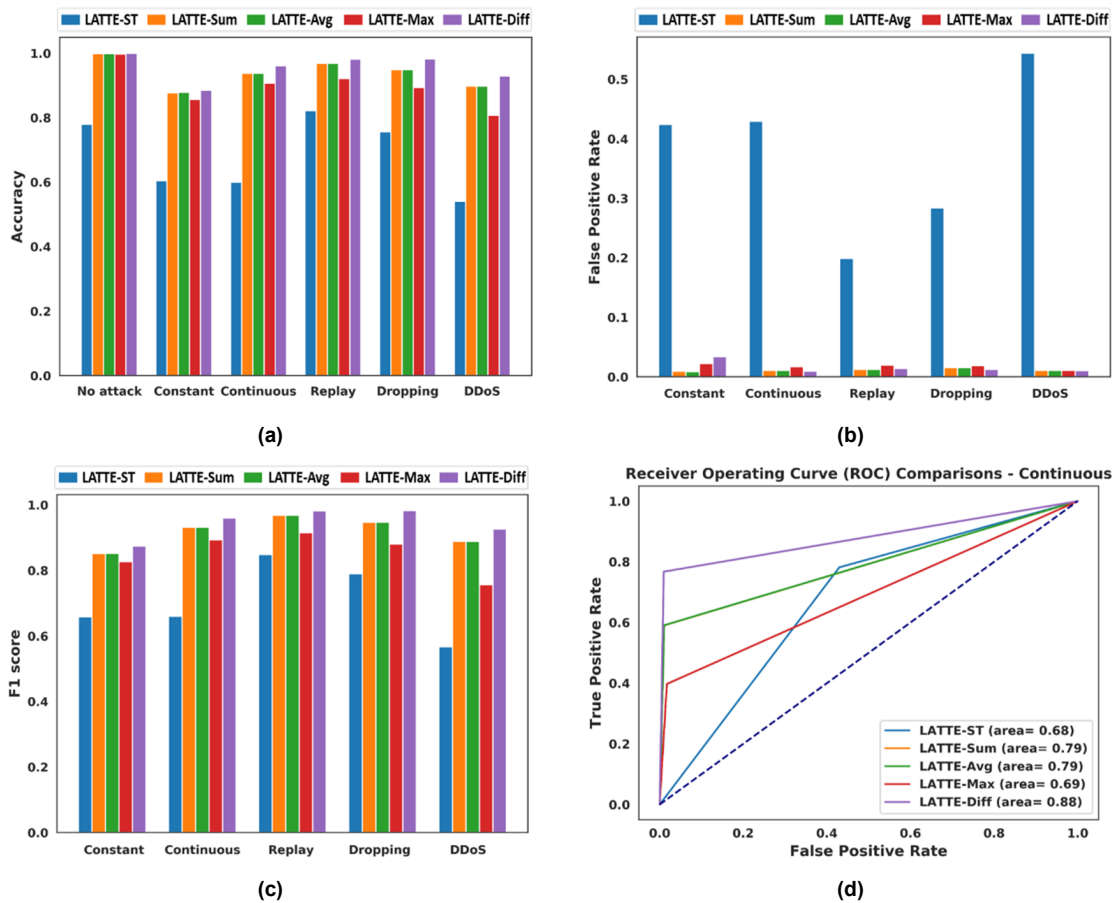

Figure 8: Comparison of (a) detection accuracy, (b) false-positive rates, (c) F1 score of *LATTE* variants under different attack scenarios, and (d) ROC curve with AUC for continuous attack.

## 6.3 Comparison with Prior Works

We compared our *LATTE* framework with BWMP [30], HAbAD [34], a variant of HAbAD called S-HAbAD [34], and RepNet [28]. Fig 9(a)-(c) show the detection accuracy, false-positive rate, and F1 score respectively for these frameworks under different attack scenarios. It can be observed that *LATTE* outperforms all the prior works in terms of detection accuracy, false-positive rate, and F1 score. This is due to three factors. Firstly, the stacked LSTM encoder-decoder structure provides adequate depth to the model to learn complex time-series patterns. This can be seen when comparing HAbAD with S-HAbAD, as the latter differs only in terms of stacked LSTM layers in comparison to the former. Second, the self-attention mechanism helps *LATTE* in learning message sequences that have very long-term dependencies. Lastly, the use of powerful OCSVMs as non-linear classifiers helps in constructing a highly efficient classifier. These factors together resulted in the superior performance of *LATTE* compared to all the comparison works. On average, across all attacks, *LATTE* was able to achieve an average of 18.94% improvement in accuracy, 19.5% improvement in F1 score, 37% improvement in AUC and 79% reduction in false positive rate, and up to 47.8% improvement in accuracy, 37.5% improvement in F1 score, 76% improvement in AUC and 95% reduction in false positive rate.

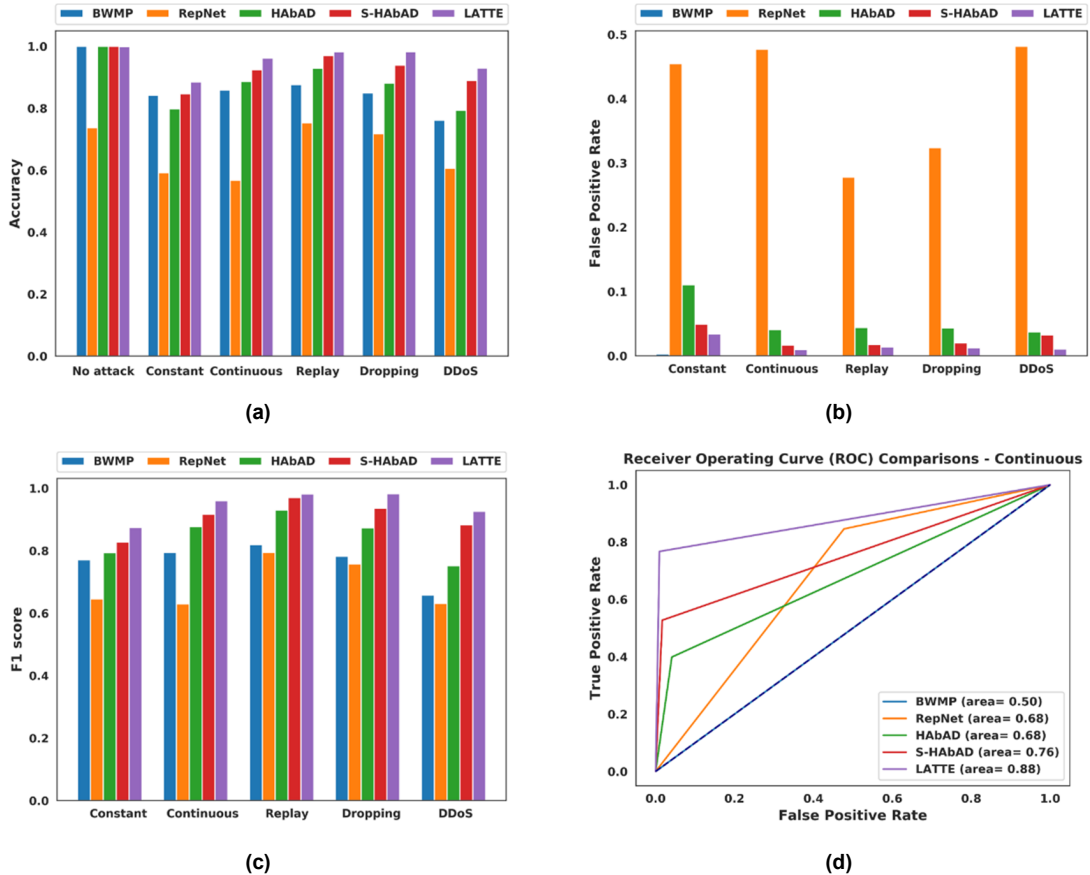

Figure 9: Comparison of (a) accuracy, (b) false-positive rates, (c) F1 score of *LATTE* and the comparison works under different attack scenarios, and (c) ROC curve with AUC for continuous attack.

To highlight the effectiveness of our proposed LATTE framework, we further compared LATTE with statistical and proximity based techniques. We selected Bollinger bands (a popular statistical technique used in the finance domain) as the candidate for a statistical technique to detect anomalies in time series data. Bollinger bands generate envelopes that are two standard deviation levels above and below the moving average. In this work, we considered two different moving average based variants of the approach: (i) simple moving average (SMA), and (ii) exponential weighted moving average (EWMA) similar to [42]. We also compared LATTE against a local outlier factor (LOF) [43] based anomaly detection technique, which is a popular proximity-based anomaly detection technique. The LOF algorithm measures the local deviation of each point in the dataset with respect to the neighbors (given by KNN) to detect anomalies. The F1 score results for SMA based Bollinger bands (SMA-BB), EWMA based Bollinger bands (EWMA-BB), LOF, and LATTE under different attack scenarios are shown in Fig. 10. It can be seen that LATTE outperforms both statistical and proximity-based anomaly detection techniques under different attack scenarios. This is mainly because the complex patterns in CAN message data are hard to capture using statistical and proximity-based techniques. On the other hand, the LSTM based predictor model in our proposed LATTE framework learns these complex patterns and is thus able to more efficiently detect complex attacks.

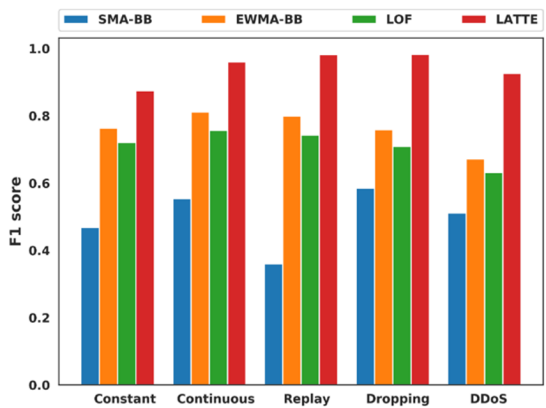

Figure 10: Comparison of F1 score for SMA-BB [42], EWMA-BB [42], LOF [43], and *LATTE* under different attack scenarios.

## 6.4 Overhead Analysis

In this subsection, we present an overhead analysis of our *LATTE* framework. We quantify the overhead of our *LATTE* framework and the comparison works using memory footprint, the number of model parameters, and the inference time metrics. We profiled each framework on a dual core ARM Cortex- A57 CPU on an NVidia Jetson TX2 board (shown in Fig. 11), which has similar specifications to that of a real-world ECU. We repeated the inference time experiment 10 times and computed the average inference time. Moreover, in this work, we consider a total buffer size of 2.25 KB. This accounts for the storage of 32 CAN message payloads (0.25 KB assuming a worst case max payload of 8 Bytes) that represent the subsequence length number of past messages, and storage of 16 CAN message frames (2 KB assuming the CAN extended protocol and a worst case max payload of 8 Bytes) that is used by the transceiver. In this work, we only introduce the additional 0.25 KB storage as the 2 KB transceiver buffer space is already available in the traditional CAN communication controller interfaces. We consider a 2 KB transceiver buffer, as it is the most commonly used size in many real-world automotive ECUs such as Woodward SECM 112, and dSpace MicroAutoBox. Additionally, we computed the area overhead of the 0.25 KB buffer using CACTI tool [44] by modeling the buffer as a scratchpad cache using 32 nm technology node. Our additional 0.25 KB buffer resulted in a minimal area overhead of around 581.25 µm$^2$. From Table 2, we can observe that our *LATTE* framework has minimal overhead compared to both attention-based prior works (HAbAD and S-HAbAD) and the non-attention based work (BWMP except RepNet). The high runtime and memory overhead in HAbAD and S-HAbAD is associated with the use of KNNs. KNN does not generalize the data in advance, but rather scans through each training data sample to make a prediction. This makes it very slow and consume high memory overhead (due to the requirement of having training data available at runtime). It needs to be noted that, even though RepNet has the lowest memory and runtime overhead, it fails to capture the complex attack patterns due to the smaller model size and the lack of ability of simple RNNs to learn long-term dependencies, leading to poor performance (as shown in Fig. 9).

Table 2: Overhead of *LATTE*, BWMP [30], HAbAD [34], S-HAbAD [34], RepNet [28]

| Framework | Memory footprint (KB) | #Model parameters (x$10^3$) | Average inference time (µs) |
|---|---|---|---|
| BWMP [30] | 13,147 | 3435 | 644.76 |
| HAbAD [34] | 4558 | 64 | 685.05 |
| S-HAbAD [34] | 5600 | 325 | 976.65 |
| *LATTE* | 1439 | 331 | 193.90 |
| RepNet [28] | 5 | 0.8 | 68.75 |

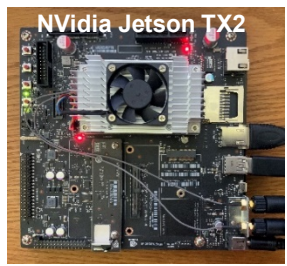

Figure 11: Nvidia Jetson TX2 development board

Assuming a distributed anomaly detection implementation, we factor in this additional latency into our real-time constraints for message transmission (i.e., a constant time overhead). But since the latency overhead (shown in Table 2) is very minimal, we envision that our proposed *LATTE* framework will have a minimal change in the timing constraints when compared to the prior works. Moreover, the deadline constraints for some of the fastest (i.e., most stringent) safety-critical applications are around 10 ms, which is much higher than our overhead that is around 193 µs. Hence, the additional latency due to our anomaly detection should not violate any safety-critical deadlines. In summary, from Fig. 9 and the results in Table 2, we can clearly observe that *LATTE* achieves superior performance compared to all of the comparison works across diverse attack scenarios, while maintaining relatively low memory and runtime overhead.

## 7 CONCLUSION

In this paper, we proposed a novel stacked LSTM with self-attention framework called *LATTE* that learns the normal system behavior by learning to predict the next message instance under normal operating conditions. We presented a one class support vector (OCSVM) based detector model to detect cyber-attacks by monitoring the message deviations from the normal behavior. We presented a detailed analysis by comparing our proposed model with multiple variants of our model and the best-known prior works in this area. Our *LATTE* framework surpasses all the variants and the best-known prior works under different attack scenarios while having a relatively low memory and runtime overhead. As a part of future work, we will explore extending our framework to detect malfunctions such as blockages, deadlocks, and faults in addition to detecting malicious behavior on the in-vehicle network.